# A Fully Progressive Approach to Single-Image Super-Resolution

Yifan Wang[1,2]   Federico Perazzi[2]   Brian McWilliams[2]
Alexander Sorkine-Hornung[2]   Olga Sorkine-Hornung[1]   Christopher Schroers[2]

[1]ETH Zurich   [2]Disney Research

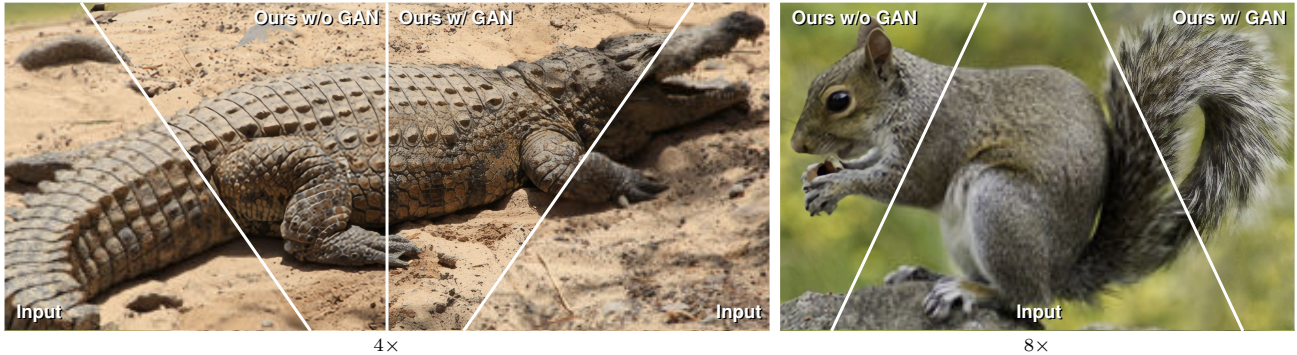

Figure 1: Examples of our 4× and 8× upsampling results. Our model without GAN sets a new state-of-the-art benchmark in terms of PSNR/SSIM; our GAN-extended model yields high perceptual quality and is able to hallucinate plausible details up to 8× upsampling ratio.

## Abstract

*Recent deep learning approaches to single image super-resolution have achieved impressive results in terms of traditional error measures and perceptual quality. However, in each case it remains challenging to achieve high quality results for large upsampling factors. To this end, we propose a method (**ProSR**) that is progressive both in architecture and training: the network upsamples an image in intermediate steps, while the learning process is organized from easy to hard, as is done in curriculum learning. To obtain more photorealistic results, we design a generative adversarial network (GAN), named **ProGanSR**, that follows the same progressive multi-scale design principle. This not only allows to scale well to high upsampling factors (e.g., 8×) but constitutes a principled multi-scale approach that increases the reconstruction quality for all upsampling factors simultaneously. In particular ProSR ranks 2nd in terms of SSIM and 4th in terms of PSNR in the NTIRE2018 SISR challenge [34]. Compared to the top-ranking team, our model is marginally lower, but runs 5 times faster.*

## 1. Introduction

The widespread availability of high resolution displays and rapid advancements in deep learning based image processing has recently sparked increased interest in super-resolution. In particular, approaches to single image super resolution (SISR) have achieved impressive results by learning the mapping from low-resolution (LR) to high-resolution (HR) images based on data. Typically, the upscaling function is a deep neural network (DNN) that is trained in a fully supervised manner with tuples of LR patches and corresponding HR targets. DNNs are able to learn abstract feature representations in the input image that allow some degree of disambiguation of the fine details in the HR output.

Most existing SISR networks adopt one of the two following *direct* approaches. The first upsamples the LR image with a simple interpolation method (e.g., bicubic) in the beginning and then essentially learns how to deblur [7, 20, 33]. The second proposes upsampling only at the end of the processing pipeline, typically using a sub-pixel convolution layer [30] or transposed convolution layer to recover the HR result [8, 23, 30, 36]. While the first class of approaches has a large memory footprint and a high computational cost, as it operates on upsampled images, the second class is more prone to checkerboard artifacts [27] due to simple concatenation of upsampling layers. Thus it remains challenging to achieve high quality results for large upsampling factors.

In this paper, we propose a method that is progressive both in architecture and training. We design the network to reconstruct a high resolution image in intermediate steps by progressively performing a 2× upsampling of the input from the previous level. As building blocks for each level of the pyramid, we propose *dense compression units*, which are adapted from dense blocks [16] to suit super-resolution.

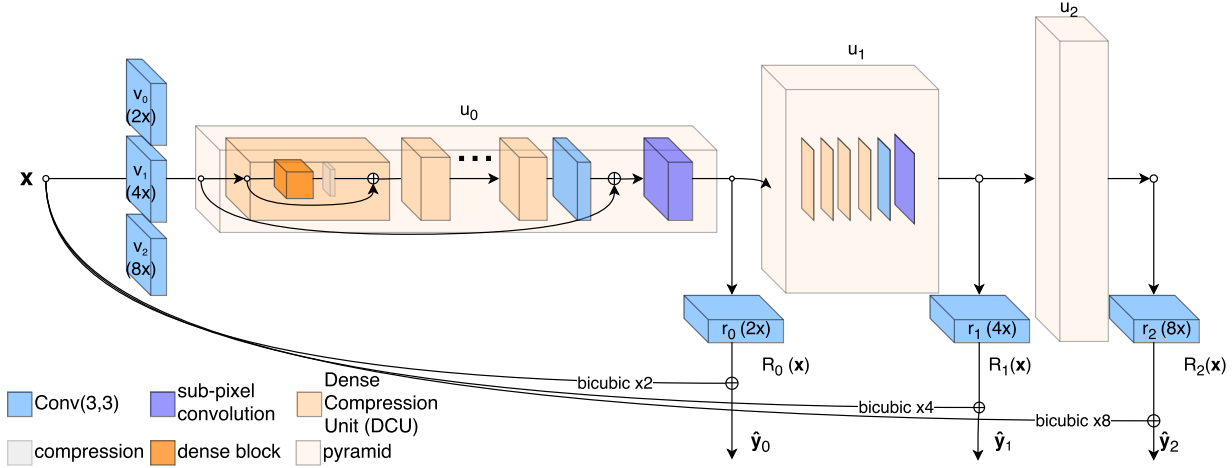

Figure 2: Asymmetric pyramidal architecture. More DCUs are allocated in the lower pyramid level to improve the reconstruction accuracy and to reduce memory consumption.

Compared to existing progressive SISR models [21, 22], we improve the reconstruction accuracy by simplifying the information propagation within the network; furthermore we propose to use an *asymmetric pyramidal* structure with more layers in the lower levels to enable high upsampling ratios while remaining efficient. To obtain more photorealistic results, we adopt the GAN framework [14] and design a discriminator that matches the progressive nature of our generator network by operating on the residual outputs of each scale. Such paired progressive design allows us to obtain a *multi-scale* generator with a unified discriminator in a single training.

In this framework, we can naturally utilize a form of *curriculum learning*, which is known to improve training [4] by organizing the learning process from easy (small upsampling factors) to hard (large upsampling factors). Compared to common multi-scale training, the proposed training strategy not only improves results for all upsampling factors, but also significantly shortens the total training time and stabilizes the GAN training.

We evaluate our progressive multi-scale approach against the state-of-art on a variety of datasets, where we demonstrate improved performance in terms of traditional error measures (e.g., PSNR) as well as perceptual quality, particularly for larger upsampling ratios.

## 2. Related Work

Single image super-resolution techniques (SISR) have been an active area of investigation for more than a decade [12]. The ill-posed nature of this problem has typically been tackled using statistical techniques: most notably image priors such as heavy-tailed gradient distributions [10, 29], gradient profiles [32], multi-scale recurrence [13], self-examples [11], and total variation [26]. In contrast, exemplar-based approaches such as nearest-neighbor [12] and sparse dictionary learning [35, 37, 39] have exploited the inherent redundancy of large-scale image datasets. Recently, Dong *et al*. [6] showed the superiority of a simple three-layer convolutional network (CNN) over sparse coding techniques. Since then, deep convolutional architectures have consistently pushed the state-of-art forward.

**Direct vs. Progressive Reconstruction.** Direct reconstruction techniques [7, 20, 23, 24, 33, 36] upscale the image to the desired spatial resolution in a single step. Early approaches [7, 20, 33] upscale the LR image in a preprocessing step. Thus, the CNN learns to deblur the input image. However, this requires the network to learn a feature representation for a high-resolution image which is computationally expensive [30]. To overcome this limitation, many approaches opt for operating on the low dimensional features and perform upsampling at the end of the network via sub-pixel convolution [30] or transposed convolution.

A popular progressive reconstruction approach is described by LapSRN by Lai *et al*. [21]. In their work, the upsampling follows the principle of Laplacian pyramids, i.e. each level learns to predict a residual that should explain the difference between a simple upscale of the previous level and the desired result. Since the loss functions are computed at each scale, this provides a form of intermediate supervision. Lai *et al*. improved their method with deep and wider recursive architecture and multi-scale training [22]. While [22] improved the accuracy, there remains a considerable gap between the top-performing approach in terms of PSNR [24]. In particular, as we show in Section 4.2, the Laplacian pyramidal structure aggravates the optimization difficulty. Furthermore, the recursive pyramids result in quadratic growth of computation in the higher pyramid level, becoming the bottleneck for reducing runtime and expanding the network capability. Lastly, in addition to a progressive generator, we also propose a progressive discriminator along with a progressive training strategy.

**Perceptual Loss Functions.** The aforementioned techniques optimize the reconstruction error by minimizing the $\ell_1$-norm and descendants such as the Charbonnier penalty function [21]. Although these approaches yield small reconstruction errors, they are unable to hallucinate perceptually plausible high-frequencies details. To this end, Ledig *et al.* [23] proposed a perceptual loss function consisting of a content loss that captures perceptual similarities and an adversary to steer the reconstruction closer to the latent manifold of natural solutions. Based on this, Sajjadi *et al.* [28] apply an additional texture loss to encourage similarity with the original image. In contrast to these works, we design a discriminator that operates on the residual outputs of each scale and train progressively with a strategy based on curriculum learning. With this, our GAN model is able to upsample perceptually pleasing SR images for *multiple scales* up to 8×.

## 3. Progressive Multi-scale Super-resolution

Given a set of $n$ LR input images with corresponding HR target images $\{(\boldsymbol{x}_1, \boldsymbol{y}_1), \ldots, (\boldsymbol{x}_n, \boldsymbol{y}_n)\}$, we consider the problem of estimating an upscaling function $u : X \rightarrow Y$, where $X$ and $Y$ denote the space of LR and HR images, respectively. Finding a suitable parameterisation for the upscaling function $u$ for large upsampling ratios, is challenging: the larger the ratio, the more complex the function class required.

To this end, we propose a progressive solution to learn the upscaling function $u$. In the following, we propose our pyramidal architecture, **ProSR**, for multi-scale super-resolution in Section 3.1 and 3.2. In Section 3.3 we propose **ProGanSR**, a progressive multi-scale GAN for perceptual enhancement. Finally, we discuss a curriculum learning scheme in Section 3.4.

### 3.1. Pyramidal Decomposition

We propose a *pyramidal* decomposition of $u$ into a series of simpler functions $u_0, \ldots, u_s$. Each function—or level— is tasked with refining the feature representation and performing a 2× upsampling of its own input. Each level of the pyramid consists of a cascade of *dense compression units* (DCUs) followed by a sub-pixel convolution layer. We assign more DCUs in the lower pyramid levels, resulting in the *asymmetric structure*. Having more computation power in the lower pyramid not only reduces the memory consumption but also increases the receptive field with respect to the original image, hence it outperforms the symmetric variant in terms of reconstruction quality and runtime. While the decomposition of $u$ is shared among the pyramid levels, we also use two scale-specific sub-networks, denoted by $v_s$ and $r_s$, which allow for an individual transformation between scale-varying image space and a normalized feature space. A schematic illustration of our *progressive* upsampling architecture is detailed in Figure 2.

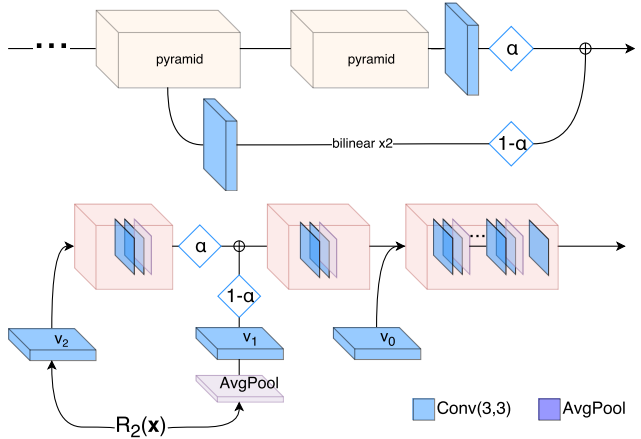

Figure 3: Schematic illustration of the blending procedure in curriculum training for the generator (top) and the discriminator (bottom).

To simplify learning, the network is designed to output the residual

$$\mathcal{R}_s(\boldsymbol{x}) = (r_s \circ u_s \circ \cdots \circ u_0 \circ v_s)(\boldsymbol{x}) \quad (1)$$

w.r.t a fixed upsampling of the input $\varphi_s(\boldsymbol{x})$ through *e.g.* bicubic interpolation. Thus, for a given scaling factor $s$ the estimated HR image can be computed as

$$\hat{\boldsymbol{y}} = \mathcal{R}_s(\boldsymbol{x}) + \varphi_s(\boldsymbol{x}). \quad (2)$$

Notably, our network doesn't follow the Laplacian pyramid principle like in [21, 22], i.e. the intermediate sub-net outputs are neither supervised nor used as base image in the subsequent level. Such design performs favorably over the Laplacian alternative, as it simplifies the backward-pass and thus reduces the optimization difficulty. Additionally we do not downsample the groundtruth to create labels, which is done for the intermediate supervision in [21, 22]. This avoids artefacts that may result from subsampling.

### 3.2. Dense Compression Units

We base the construction of each pyramid level on the recently proposed DenseNet architecture [16]. Similarly to skip connections [15], dense connections improve gradient flow alleviating vanishing and shattered gradients [3].

The core component in each level of the pyramid is a dense compression unit (DCU), which consists of a modified densely connected block followed by $1 \times 1$ convolution CONV(1,1).

The original dense layer is composed of BN-RELU-CONV(1,1)-BN-RELU-CONV(3,3). Following recent practice in super-resolution [9, 24, 38], we remove all batch normalizations. However, since the features from previous layers may have varying scales, we also remove the first ReLU to rescale the features with CONV(1-1). This leads to a modified dense layer composition: CONV(1,1)-RELU-CONV(3,3).

Contrary to DenseNet, we break the dense connection at the end of each DCU with a CONV(1,1) compression layer, which re-assembles information efficiently and leads to a slight performance gain in spite of the breakage of dense connection. For a very deep model we apply pyramid-wise as well as local residual links to improve the gradient propagation as shown in Figure 2.

### 3.3. Progressive GAN

Generative adversarial networks (GANs) [14] have emerged as a powerful method to enhance the perceptual quality of the upsampled images [14, 23, 28] in SISR.

However, training GANs is notoriously difficult and success at applying GANs to SISR has been limited to single-scale upsampling at relatively low target resolutions. In order to enable multi-scale GAN-enhanced SISR, we propose a modular and progressive discriminator network similar to the generator network proposed in the previous section. As Figure 3 illustrates the architecture has a reverse pyramid structure, where each level gradually reduces the spatial dimension of the input image with AVGPOOLING. To accommodate the multi-scale outputs from the generator, the network is fully convolutional and outputs a small patch of features similar to PatchGAN [18]. The complete specs of the discriminator can be found in the supplemental material.

Similar to the generator network, the discriminator operates on the residual between the original and bicubic upsampled image. This allows both generator and discriminator to concentrate only on the important sources of variation which are not already well captured by the standard upsampling operation. Since these regions are challenging to upsample well, they correspond to the largest perceptual errors. This can also be viewed as subtracting a data-dependent baseline from the discriminator which helps to reduce variance.

As the training objective, we use the more stable least square loss instead of the original cross-entropy loss [25]. Denoting the predicted residual and real residual as $\hat{r}$ and $r$, the discriminator loss and generator loss for a training example of scale $s$ can be expressed as

$$\mathcal{L}_{\mathcal{D}_s}^i = (\mathcal{D}(\hat{r}_i^s))^2 + (\mathcal{D}(r_i^s) - 1)^2 \quad (3)$$

$$\mathcal{L}_{\mathcal{R}_s}^i = (\mathcal{D}(\hat{r}_i^s) - 1)^2 + \quad (4)$$
$$\sum_{k \in \{2,4\}} \|\Phi_k(\hat{y}_i) - \Phi_k(y_i)\|^2 ,$$

where $\Phi_k$ denotes the $k$-th pooling layer input in VGG16 [31].

### 3.4. Curriculum Learning

Curriculum learning [4] is a strategy to improve training by gradually increasing the difficulty of the learning task. It is often used in sequence prediction tasks and in sequential decision making problems where large speedups in training time and improvements in generalisation performance can be obtained.

The pyramidal decomposition of $u$ allows us to apply curriculum learning in a natural way. The loss for a training example $(x_i^s, y_i)$ of scale $s$ can be defined as

$$\mathcal{L}_{\mathcal{R}_s}^i = \|\mathcal{R}_s(x_i^s) + \varphi_s(x_i^s) - y_i\|_1 \quad (5)$$

where $x_i^s$ corresponds to $s\times$ downsampled version of $y_i$. Then the goal at scale $s$ is to find

$$\hat{\theta}_s = \underset{\theta_s}{\operatorname{argmin}} \sum_{s' \leq s} \sum_i \mathcal{L}_{\mathcal{R}_{s'}}^i , \quad (6)$$

where $\theta_s$ parameterises all functions in and below the current scale $(u_0, v_0, r_0, \ldots, u_s, v_s, r_s)$ according to our pyramidal network shown in Figure 2. Our training curriculum starts by training only the $2\times$ portion of the network. When we proceed to a new phase in the curriculum (e.g. to $4\times$), a new level of the pyramid is gradually blended in to reduce the impact on the previously trained layers. As Figure 3 shows, this is achieved by linearly combining the output of the newly added level with the upsampled output of the previous level. A similar idea was proposed in [19]. As a result we incrementally add training pairs of the next scale. Finally, to assemble the batches, we randomly select one of the scales $s$ to avoid mixing batch statistics as suggested in [2].

Compared to simple multi-scale training where training examples from different scales are simultaneously fed to the network, such progressive training strategy greatly shortens the total training time. Furthermore, it yields a further performance gain for all included scales compared to single-scale and simple multi-scale training and alleviates instabilities in GAN training.

## 4. Evaluation

Before we compare with popular state-of-the-art approaches, we first discuss the benefits of each of our proposed components using a small 24-layer model.

All presented models are trained with the DIV2K [34] training set, which contains 800 high-resolution images. The training details are listed in the supplemental material. For evaluation, the benchmark datasets Set5 [5], Set14 [40], BSD100 [1], Urban100 [17], and the DIV2K validation set [34] are used. As it is commonly done in SISR, all evaluations are conducted on the luminance channel.

### 4.1. Ablation study

| Ablation Study | Method | PSNR | Parameters |
|---|---|---|---|
| Baseline | Single Dense Block, Dense Layer BRCBRC, Single Scale | 28.30 | 8.22M |
| Block Division | 4 DCUs | 28.32 | 1.79M |
| Architecture | Asymmetric Pyramid | 28.41 | 1.89M |
| Training | Curriculum Learning | 28.45 | 1.89M |
| Very Deep Model | Increased network width and depth longer training | 28.94 | 13.4M |

Table 1: Overview of experiments in the ablation study. The introduction of DCUs, block division, an asymmetric pyramid layout, and curriculum learning allow to consistently increase reconstruction quality. Reported values refer to $4\times$ results of Set14.

Table 1 summarizes the consistent increase in reconstruction quality stemming from each proposed component. As a baseline, we start from a single dense block with two sub-pixel upsampling layers in the end and a residual connection from the LR input to the final output. In the following, we describe the individual steps in more detail.

**Dense Compression Units.** To demonstrate the benefit of DCUs described in Section 3.2, we replace the single-block from the baseline model with multiple DCUs. As Table 1 shows, the number of network parameters can be drastically reduced without harming the reconstruction accuracy. We can even observe a slight performance gain as the network is able to reassemble features more efficiently due to the injection of compression layers.

**Asymmetric Pyramid.** In this section we show the advantage of the proposed asymmetric pyramidal architecture. We compare the following constellations while keeping the total number of DCUs constant:

| Model | Architecture |
|---|---|
| Direct | $D - D - D - D - S - S$ |
| Asymmetric Pyramid | $D - D - D - S - D - S$ |

Here, $D$ denotes a dense compression unit with 6 dense layers and $S$ denotes the sub-pixel upsampler. As Table 1 shows, the asymmetric pyramidal architecture considerably improves the reconstruction accuracy compared to direct upsampling. This demonstrates the advantage of utilizing high-dimensional features directly.

**Curriculum Learning.** We extend the 4-DCU asymmetric pyramid model to $8\times$ upsampling to quantify the benefit of curriculum learning over simultaneous multi-scale training. As Table 2 shows, simultaneous training typically has small or even negative impact on the lowest scale ($2\times$), which is also evident in VDSR [20] (see Table 2). On the other hand, curriculum learning always improves the reconstruction quality and outperforms simultaneous training by an average of 0.04dB.

Furthermore, curriculum learning considerably shortens the training time. As Figure 4 shows, the network reaches the same number of epochs and quality faster than simultaneous training, since the $2\times$ subnet requires less computation and hence less time for each update.

### 4.2. Comparison with other progressive architectures.

In contrast to our approach, existing progressive methods [21, 22] typically rely on deep supervision. They impose a loss on all scales which can be denoted as

$$\mathcal{L}_s^i = \sum_{s'<s} \ell_1\left(\psi_{s'}(\boldsymbol{y}_i), \hat{\boldsymbol{y}}_i^{s'}\right) + \ell_1\left(\boldsymbol{y}_i^s, \hat{\boldsymbol{y}}_i^s\right), \quad (7)$$

with $\psi_{s'}$ being a downsampling operation to scale $s'$. Futhermore, following the structure of a Laplacian pyramid, each level is encouraged to learn the difference between a bicubic upscale of the previous level instead of the upsampled LR image. Thus the residual connections are given by

$$\hat{\boldsymbol{y}}^s = \hat{\boldsymbol{r}}^s + \varphi_2\left(\boldsymbol{y}^{s-1}\right), \quad (8)$$

where $\varphi_2$ denotes an upscaling operator by a factor of 2.

We also evaluate such alternative progressive architecture but observed large decrease in PSNR as shown in Table 3. Therefore, we conclude that it is less stable to use varying sub-scale upsampling results as base images compared to fixed interpolated results and that using a downsampling kernel to create the HR label images could introduce undesired artefacts.

### 4.3. Comparison with State-of-the-art Approaches

In this section, we provide an extensive quantitative and qualitative comparison with other state-of-the-art approaches.

**Quantitative Comparison.** For a quantitative comparison, we benchmark against VDSR [20], DRRN [33], LapSRN [21], MsLapSRN [22], EDSR. We obtained models from Lai *et al.* [22] for $8\times$ versions of VDSR and DRRN, that have been retrained with 8× data. To produce 8× EDSR results, we extend their 4× model by adding another sub-pixel convolution layer. For training, we follow their practice which means we initialize the weights of the 8× model from the pretrained 4× model.

| Improvement w.r.t single-scale 2×/4×/8× (dB) | Set5 | Set14 | B100 | U100 | DIV2K | average |
|---|---|---|---|---|---|---|
| simultaneous | -0.05/+0.09/-0.01 | +0.01/+0.03/+0.05 | +0.02/+0.01/+0.03 | +0.12/+0.06/+0.08 | +0.06/-0.02/+0.05 | +0.06/+0.02/+0.05 |
| curriculum | +0.05/+0.11/+0.08 | +0.08/+0.04/+0.06 | +0.07/+0.03/+0.05 | +0.21/+0.09/+0.08 | +0.13/+0.02/+0.05 | +0.13/+0.05/+0.06 |

Table 2: Gain of simultaneous training and curriculum learning w.r.t. single-scale training on all datasets. The average is computed accounting the number of images in the datasets. Curriculum learning improves the training for all scales while simultaneous training hampers the training of the lowest scale.

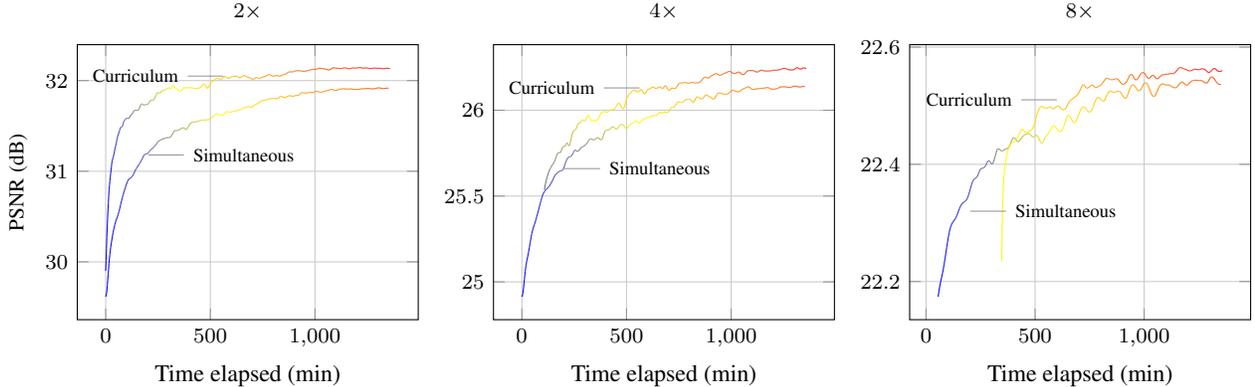

Figure 4: Training time comparison between curriculum learning and multiscale simultaneous learning. We train the multiscale model and plot the PSNR evaluation of the individual scales. The elapsed epoch is encoded as the line color. Because curriculum learning activates the smaller subnets first, it requires much less time to reach the same evaluation quality.

| model | | B100 2× | B100 4× | B100 8× | Set14 2× | Set14 4× | Set14 8× |
|---|---|---|---|---|---|---|---|
| single scale | ours | - | 27.44 | - | - | 28.41 | - |
| | alt | - | 27.32 | - | - | 28.20 | - |
| multi scale | ours | 31.95 | 27.47 | 24.75 | 33.24 | 28.45 | 24.86 |
| | alt | 31.92 | 27.38 | 24.70 | 33.22 | 28.28 | 24.76 |

Table 3: Comparison with other progressive approaches.

Due to discrepancy in the model size within existing approaches, we divide them into two classes based on whether they have more or less than 5 million parameters. Accordingly, we provide two models with different sizes, denoted as ProSR$_s$ and ProSR$_\ell$, to compete in both classes. ProSR$_s$ has 56 dense layers in total with growth-rate $k = 12$ and a total of 3.1M parameters. ProSR$_\ell$ has 104 dense layers with growth-rate $k = 40$ and 15.5M parameters which is roughly a third of the parameters of EDSR.

Table 4 summarizes the quantitative comparison with other state-of-the-art approaches in terms of PSNR. An extended list that includes SSIM scores can be found in the supplemental material. As Table 4 shows, ProSR$_s$ achieves the lowest error in most datasets. The very deep model, ProSR$_\ell$, shows consistent advantage in higher upsampling ratios and is comparable with EDSR in 2×. In general, our progressive design allows to raise the margin in PSNR between our results and the state-of-the art as the upsampling ratio increases.

**Qualitative comparison.** First, we qualitatively compare our method without GAN to other methods that also minimise the $\ell_1$ loss or related norms. Figure 7 show results of our method and the most recent state-of-the-art approaches in 4× and 8×.

Concerning our perceptually-driven model with GAN, we compare with SRGAN [23] and EnhanceNet [28]. As Figure 5 shows, the hallucinated details align well with fine structures in the ground truth, even though we do not have an explicit texture matching loss as EnhanceNet [28]. While SRGAN and EnhanceNet can only upscale 4×, our method is able to extend to 8×. Results are shown in Figure 6. We provide an extended qualitative comparison in the supplemental material.

## 5. Runtime.

The asymmetric pyramid architecture contributes to faster runtime compared to other approaches that have sim-ilar reconstruction accuracy. In our test environment with NVIDIA TITAN XP and cudnn6.0, ProSR$_\ell$ takes on average 0.8s, 2.1s and 4.4s to upsample a $520 \times 520$ image by $2\times$, $4\times$ and $8\times$. In the NTIRE challenge, we reported the runtime including geometric ensemble, which requires 8 forward passes for each transformed version of the input image. Nonetheless, our runtime is still 5 times faster than the top-ranking team.

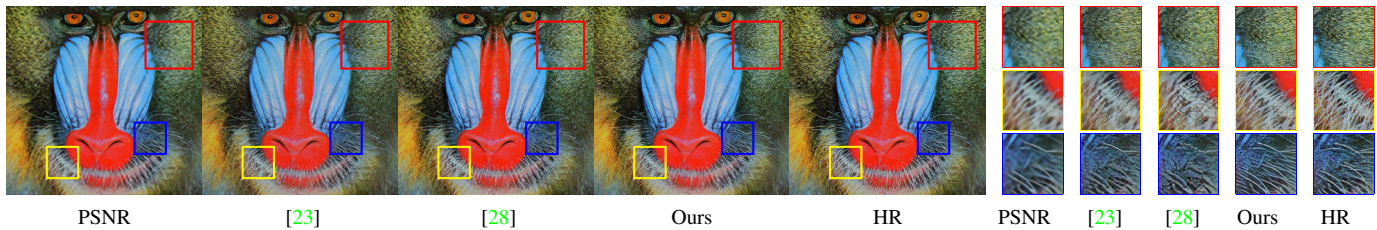

Figure 5: Comparison of $4\times$ GAN results (best viewed when zoomed in). Our approach is less prone to artefacts and aligns well with the original image.

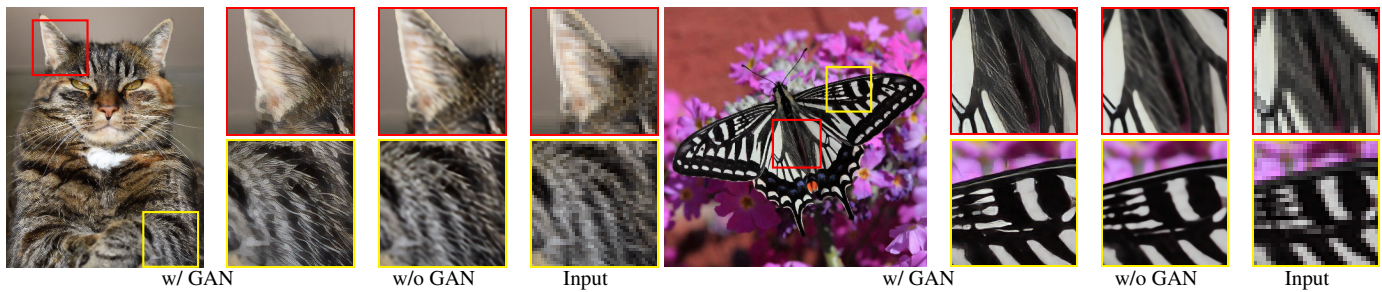

Figure 6: Hallucinated details in $8\times$ upsample result with adversarial loss.

## 6. NTIRE Challenge

The "New Trends in Image Restoration and Enhancement" (NTIRE) 2018 super-resolution challenge aims at benchmarking SISR methods in challenging scenarios.

In particular, one of the challenge tracks targets $8\times$ upscaling, where the low resolution images are generated with known downsampling kernels (bicubic). We participated in the challenge with the ProSR$_\ell$ network. In addition to the method described above, we utilised the geometry ensemble used in [24], which yielded a 0.07dB PSNR gain in the validation set. Our model ranks 2nd in terms of SSIM and 4th in terms of PSNR. Compared to the top-ranking team, our model is marginally lower by 0.002 and 0.04dB in SSIM and PSNR respectively, but runs 5 times as fast in test time.

Other tracks in the challenge target $4\times$ upscaling but consider unknown degradation. Given that this task is different to the bicubic $8\times$ setting, the participating teams and the rankings differ. Without specific adaptation for this scenario, we also participated in these tracks for completeness and ranked in the mid-range (7th/9th/7th). We believe further improvement can be achieved with targeted preprocessing and extended training data.

## 7. Conclusion

In this work we propose a progressive approach to address SISR. Specifically, we design a novel architecture **ProSR** that incrementally upsamples the input image by a factor of 2, forming a pyramidal shape. In particular, we show the benefit of an asymmetric pyramid design with more computation resource in lower pyramid levels both in terms of reconstruction quality and the memory efficiency. To allow for a very deep architecture and thus more modeling power, we have proposed Dense Compression Units (DCU) as building blocks which leverage dense connectivity. Furthermore we leverage a form of curriculum learning which not only enables high-quality reconstruction in large upsampling ratios but also simultaneously increases the performance for all scales. Our **ProSR** sets a new state-of-the-art benchmark in terms of the traditional error measures and in higher upsampling ratios (e.g. $4\times$ and $8\times$) even outperforms existing methods by a large margin. Finally, we demonstrate that the same progressive principle can be applied to GAN-extended method for optimizing perceptual quality of the upsampled image. To this end we propose **ProGanSR**, which shows prominent improvement against the previous state-of-the-art both in terms of visual quality and stability; to our knowledge, it is also the first multi-scale model that could yield perceptually pleasing images in SISR up to $8\times$.

| PSNR | 2× | | | | 4× | | | | 8× | | | |
|---|---|---|---|---|---|---|---|---|---|---|---|---|
| | S14 | B100 | U100 | DIV2K | S14 | B100 | U100 | DIV2K | S14 | B100 | U100 | DIV2K |
| # params < 5M | | | | | | | | | | | | |
| VDSR | 33.05 | 31.90 | 30.77 | 35.26 | 28.02 | 27.29 | 25.18 | 29.72 | 24.26 | 24.49 | 21.70 | 26.22 |
| DRRN | 33.23 | **32.05** | 31.23 | 35.49 | 28.21 | 27.38 | 25.44 | 29.95 | 24.42 | 24.59 | 21.88 | 26.37 |
| LapSRN | 33.08 | 31.80 | 30.41 | 35.63 | 28.19 | 27.32 | 25.21 | 29.88 | 24.35 | 24.54 | 21.81 | 26.40 |
| MsLapSRN | 33.28 | **32.05** | 31.15 | 35.62 | 28.26 | 27.43 | 25.51 | **30.39** | 24.57 | 24.65 | 22.06 | 26.52 |
| SRDenseNet | - | - | - | - | 28.50 | 27.53 | **26.05** | - | - | - | - | - |
| ProSR$_s$ (ours) | **33.36** | 32.02 | **31.42** | **35.80** | **28.59** | **27.58** | 26.01 | **30.39** | **24.93** | **24.80** | **22.43** | **26.88** |
| # params > 5M | | | | | | | | | | | | |
| EDSR | 33.92 | 32.32 | **32.93** | **36.47** | 28.80 | 27.71 | 26.64 | 30.71 | 24.96 | 24.83 | 22.53 | 26.96 |
| ProSR$_l$ (ours) | **34.00** | **32.34** | 32.91 | 36.44 | **28.94** | **27.79** | **26.89** | **30.81** | **25.29** | **24.99** | **23.04** | **27.36** |

Table 4: Comparison with state-of-the-art approaches. For clarity, we highlight the best approach in **blue**.

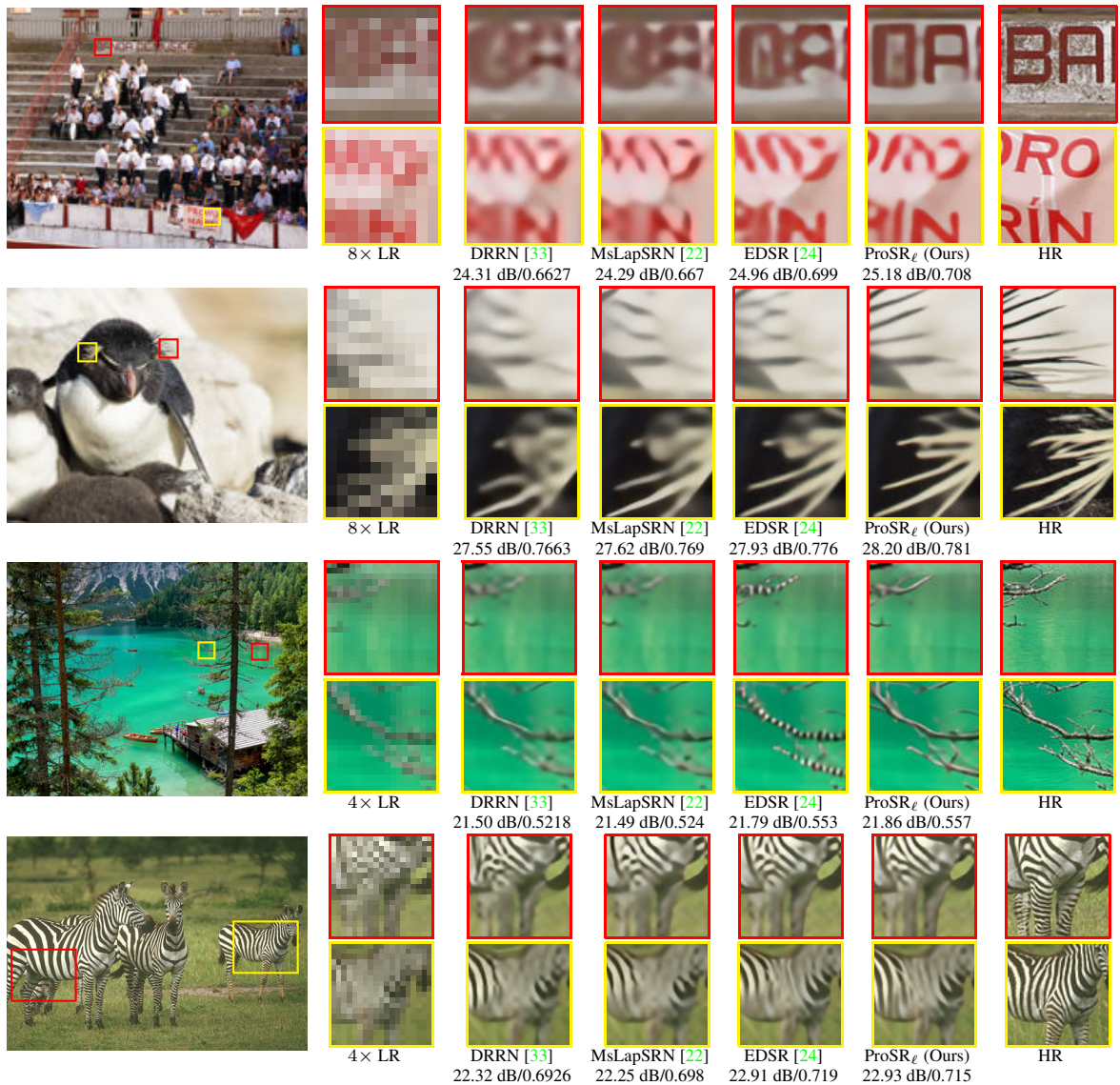

Figure 7: Visual comparison with other state-of-the-art methods.

# References

[1] P. Arbelaez, M. Maire, C. Fowlkes, and J. Malik. Contour detection and hierarchical image segmentation. *IEEE transactions on pattern analysis and machine intelligence*, 33(5):898–916, 2011. 5

[2] M. Arjovsky, S. Chintala, and L. Bottou. Wasserstein GAN. *arXiv preprint arXiv:1701.07875*, 2017. 4

[3] D. Balduzzi, M. Frean, L. Leary, J. P. Lewis, K. W.-D. Ma, and B. McWilliams. The shattered gradients problem: If resnets are the answer, then what is the question? In D. Precup and Y. W. Teh, editors, *Proceedings of the 34th International Conference on Machine Learning*, volume 70 of *Proceedings of Machine Learning Research*, pages 342–350, International Convention Centre, Sydney, Australia, 06–11 Aug 2017. PMLR. 3

[4] Y. Bengio, J. Louradour, R. Collobert, and J. Weston. Curriculum learning. In *Proceedings of the 26th annual international conference on machine learning*, pages 41–48. ACM, 2009. 2, 4

[5] M. Bevilacqua, A. Roumy, C. Guillemot, and M. L. Alberi-Morel. Low-complexity single-image super-resolution based on nonnegative neighbor embedding. 2012. 5

[6] C. Dong, C. C. Loy, K. He, and X. Tang. Learning a deep convolutional network for image super-resolution. In *Computer Vision - ECCV 2014 - 13th European Conference, Zurich, Switzerland, September 6-12, 2014, Proceedings, Part IV*, pages 184–199, 2014. 2

[7] C. Dong, C. C. Loy, K. He, and X. Tang. Image super-resolution using deep convolutional networks. *IEEE transactions on pattern analysis and machine intelligence*, 38(2):295–307, 2016. 1, 2

[8] C. Dong, C. C. Loy, and X. Tang. Accelerating the super-resolution convolutional neural network. In *European Conference on Computer Vision*, pages 391–407. Springer, 2016. 1

[9] Y. Fan, H. Shi, J. Yu, D. Liu, W. Han, H. Yu, Z. Wang, X. Wang, and T. S. Huang. Balanced two-stage residual networks for image super-resolution. In *Computer Vision and Pattern Recognition Workshops (CVPRW), 2017 IEEE Conference on*, pages 1157–1164. IEEE, 2017. 4

[10] C. Fernandez-Granda and E. J. Candès. Super-resolution via transform-invariant group-sparse regularization. In *IEEE International Conference on Computer Vision, ICCV 2013, Sydney, Australia, December 1-8, 2013*, pages 3336–3343, 2013. 2

[11] G. Freedman and R. Fattal. Image and video upscaling from local self-examples. *ACM Transactions on Graphics*, 30(2), 2011. 2

[12] W. T. Freeman, T. R. Jones, and E. C. Pasztor. Example-based super-resolution. *IEEE Computer Graphics and Applications*, 22(2):56–65, 2002. 2

[13] D. Glasner, S. Bagon, and M. Irani. Super-resolution from a single image. In *IEEE 12th International Conference on Computer Vision, ICCV 2009, Kyoto, Japan, September 27 - October 4, 2009*, pages 349–356, 2009. 2

[14] I. Goodfellow, J. Pouget-Abadie, M. Mirza, B. Xu, D. Warde-Farley, S. Ozair, A. Courville, and Y. Bengio. Generative adversarial nets. In *Advances in neural information processing systems*, pages 2672–2680, 2014. 2, 4

[15] K. He, X. Zhang, S. Ren, and J. Sun. Deep residual learning for image recognition. In *Proceedings of the IEEE conference on computer vision and pattern recognition*, pages 770–778, 2016. 3

[16] G. Huang, Z. Liu, K. Q. Weinberger, and L. van der Maaten. Densely connected convolutional networks. *arXiv preprint arXiv:1608.06993*, 2016. 1, 3

[17] J.-B. Huang, A. Singh, and N. Ahuja. Single image super-resolution from transformed self-exemplars. In *Proceedings of the IEEE Conference on Computer Vision and Pattern Recognition*, pages 5197–5206, 2015. 5

[18] P. Isola, J.-Y. Zhu, T. Zhou, and A. A. Efros. Image-to-image translation with conditional adversarial networks. *arXiv preprint arXiv:1611.07004*, 2016. 4

[19] T. Karras, T. Aila, S. Laine, and J. Lehtinen. Progressive growing of gans for improved quality, stability, and variation. *arXiv preprint arXiv:1710.10196*, 2017. 4

[20] J. Kim, J. Kwon Lee, and K. Mu Lee. Accurate image super-resolution using very deep convolutional networks. In *Proceedings of the IEEE Conference on Computer Vision and Pattern Recognition*, pages 1646–1654, 2016. 1, 2, 5

[21] W.-S. Lai, J.-B. Huang, N. Ahuja, and M.-H. Yang. Deep laplacian pyramid networks for fast and accurate super-resolution. In *IEEE Conference on Computer Vision and Pattern Recognition*, 2017. 2, 3, 5

[22] W.-S. Lai, J.-B. Huang, N. Ahuja, and M.-H. Yang. Fast and accurate image super-resolution with deep laplacian pyramid networks. *arXiv preprint arXiv:1710.01992*, 2017. 2, 3, 5, 8

[23] C. Ledig, L. Theis, F. Huszár, J. Caballero, A. Cunningham, A. Acosta, A. Aitken, A. Tejani, J. Totz, Z. Wang, et al. Photo-realistic single image super-resolution using a generative adversarial network. *arXiv preprint arXiv:1609.04802*, 2016. 1, 2, 3, 4, 6, 7

[24] B. Lim, S. Son, H. Kim, S. Nah, and K. M. Lee. Enhanced deep residual networks for single image super-resolution. In *The IEEE Conference on Computer Vision and Pattern Recognition (CVPR) Workshops*, July 2017. 2, 4, 7, 8

[25] X. Mao, Q. Li, H. Xie, R. Y. Lau, Z. Wang, and S. P. Smolley. Least squares generative adversarial networks. *arXiv preprint ArXiv:1611.04076*, 2016. 4

[26] A. Marquina and S. Osher. Image super-resolution by tv-regularization and bregman iteration. *J. Sci. Comput.*, 37(3):367–382, 2008. 2

[27] A. Odena, V. Dumoulin, and C. Olah. Deconvolution and checkerboard artifacts. *Distill*, 2016. 1

[28] M. S. M. Sajjadi, B. Schölkopf, and M. Hirsch. Enhancenet: Single image super-resolution through automated texture synthesis. *CoRR*, abs/1612.07919, 2016. 3, 4, 6, 7

[29] Q. Shan, Z. Li, J. Jia, and C. Tang. Fast image/video upsampling. *ACM Trans. Graph.*, 27(5):153:1–153:7, 2008. 2

[30] W. Shi, J. Caballero, F. Huszár, J. Totz, A. P. Aitken, R. Bishop, D. Rueckert, and Z. Wang. Real-time single image and video super-resolution using an efficient sub-pixel convolutional neural network. In *Proceedings of the IEEE Conference on Computer Vision and Pattern Recognition*, pages 1874–1883, 2016. 1, 2

[31] K. Simonyan and A. Zisserman. Very deep convolutional networks for large-scale image recognition. *arXiv preprint arXiv:1409.1556*, 2014. 4

[32] J. Sun, Z. Xu, and H. Shum. Image super-resolution using gradient profile prior. In *2008 IEEE Computer Society Conference on Computer Vision and Pattern Recognition (CVPR 2008), 24-26 June 2008, Anchorage, Alaska, USA*, 2008. 2

[33] Y. Tai, J. Yang, and X. Liu. Image super-resolution via deep recursive residual network. In *Proceedings of the IEEE Conference on Computer Vision and Pattern Recognition*, 2017. 1, 2, 5, 8

[34] R. Timofte, E. Agustsson, L. Van Gool, M.-H. Yang, L. Zhang, et al. Ntire 2017 challenge on single image super-resolution: Methods and results. In *The IEEE Conference on Computer Vision and Pattern Recognition (CVPR) Workshops*, July 2017. 1, 5

[35] R. Timofte, V. D. Smet, and L. J. V. Gool. Anchored neighborhood regression for fast example-based super-resolution. In *IEEE International Conference on Computer Vision, ICCV 2013, Sydney, Australia, December 1-8, 2013*, pages 1920–1927, 2013. 2

[36] T. Tong, G. Li, X. Liu, and Q. Gao. Image super-resolution using dense skip connections. In *The IEEE International Conference on Computer Vision (ICCV)*, Oct 2017. 1, 2

[37] J. Yang, J. Wright, T. S. Huang, and Y. Ma. Image super-resolution as sparse representation of raw image patches. In *2008 IEEE Computer Society Conference on Computer Vision and Pattern Recognition (CVPR 2008), 24-26 June 2008, Anchorage, Alaska, USA*, 2008. 2

[38] Z. Yang, K. Zhang, Y. Liang, and J. Wang. Single image super-resolution with a parameter economic residual-like convolutional neural network. In *International Conference on Multimedia Modeling*, pages 353–364. Springer, 2017. 4

[39] R. Zeyde, M. Elad, and M. Protter. On single image scale-up using sparse-representations. In *Curves and Surfaces - 7th International Conference, Avignon, France, June 24-30, 2010, Revised Selected Papers*, pages 711–730, 2010. 2

[40] R. Zeyde, M. Elad, and M. Protter. On single image scale-up using sparse-representations. In *International conference on curves and surfaces*, pages 711–730. Springer, 2010. 5